\documentclass[sigconf]{acmart}

\PassOptionsToPackage{dvipsnames}{xcolor}

\usepackage{acronym}
\usepackage{algorithm}
\usepackage{algorithmic}
\usepackage{makecell}
\usepackage[inline]{enumitem}
\usepackage{bbm}
\usepackage[skip=2pt]{caption}
\usepackage{natbib}
\usepackage{xcolor}
\usepackage{subcaption}

\renewcommand{\vec}[1]{\mathbf{#1}}

\newcommand{\header}[1]{\vspace{1mm}\noindent\textit{#1}.}

\acrodef{TCN}{Temporal Convolutional Network}
\acrodef{CNN}{Convolutional Neural Network}
\acrodef{RNN}{Recurrent Neural Network}
\acrodef{PGBM}{Probabilistic Gradient Boosting Machine}
\acrodef{GBM}{Gradient Boosting Machine}
\acrodef{GB}{Gradient Boosting}

\allowdisplaybreaks

\author{Olivier Sprangers}
\affiliation{
 \institution{AIRLab, University of Amsterdam}
 \city{Amsterdam}
 \country{The Netherlands}
}
\email{o.r.sprangers@uva.nl}
 
\author{Sebastian Schelter}
\orcid{0000-0002-1086-0202}
\affiliation{
 \institution{University of Amsterdam}
 \city{Amsterdam}
 \country{The Netherlands}
}
\email{s.schelter@uva.nl}

\author{Maarten de Rijke}
\orcid{0000-0002-1086-0202}
\affiliation{
 \institution{University of Amsterdam}
 \city{Amsterdam}
 \country{The Netherlands}
}
\email{derijke@uva.nl}

\acmSubmissionID{1763}

\copyrightyear{2021}
\acmYear{2021}
\setcopyright{acmlicensed}
\acmConference[KDD '21] {Proceedings of the 27th ACM SIGKDD Conference on Knowledge Discovery and Data Mining}{August 14--18, 2021}{Virtual Event, Singapore.}
\acmBooktitle{Proceedings of the 27th ACM SIGKDD Conference on Knowledge Discovery and Data Mining (KDD '21), August 14--18, 2021, Virtual Event, Singapore}
\acmPrice{15.00}
\acmDOI{10.1145/3447548.3467278}
\acmISBN{978-1-4503-8332-5/21/08}

\begin{document}
\fancyhead{}

\title{Probabilistic Gradient Boosting Machines \\ for Large-Scale Probabilistic Regression}

\begin{abstract}
\acp{GBM} are hugely popular for solving tabular data problems. However, practitioners are not only interested in point predictions, but also in probabilistic predictions in order to quantify the uncertainty of the predictions. Creating such probabilistic predictions is difficult with existing \ac{GBM}-based solutions: they either require training multiple models or they become too computationally expensive to be useful for large-scale settings. 

We propose \acp{PGBM}, a method to create probabilistic predictions with a single ensemble of decision trees in a computationally efficient manner. \ac{PGBM} approximates the leaf weights in a decision tree as a random variable, and approximates the mean and variance of each sample in a dataset via stochastic tree ensemble update equations. These learned moments allow us to subsequently sample from a specified distribution after training. 

We empirically demonstrate the advantages of \ac{PGBM} compared to existing state-of-the-art methods: 
\begin{enumerate*}[label=(\roman*)]
\item \ac{PGBM} enables probabilistic estimates without compromising on point performance in a single model, 
\item \ac{PGBM} learns probabilistic estimates via a single model only (and without requiring multi-parameter boosting), and thereby offers a speedup of up to several orders of magnitude over existing state-of-the-art methods on large datasets, and 
\item \ac{PGBM} achieves accurate probabilistic estimates in tasks with complex differentiable loss functions, such as hierarchical time series problems, where we observed up to 10\% improvement in point forecasting performance and up to 300\% improvement in probabilistic forecasting performance. 
\end{enumerate*}
\end{abstract}

\maketitle

\acresetall


\section{Introduction}
Forecasting practioners are increasingly interested in probabilistic forecasts instead of point forecasts, in order to obtain a notion of uncertainty of the forecast \cite{bose_probabilistic_2017}. Even though probabilistic forecasting techniques have been around for quite some time, applying these techniques in large-scale industrial settings often remains challenging. For example, retailers may require thousands of new forecasts for each product for each store. In this setting, traditional confidence interval techniques based on single model estimates are often computationally too expensive to execute on a daily basis~\cite{makridakis_m5_2020-1}. Existing \acl{GB} methods for probabilistic forecasting often require training multiple models (e.g., LightGBM~\citep{ke_lightgbm_2017} or xgboost~\citep{chen_xgboost_2016} require a separate model for each quantile of the forecast), or require computing expensive second-order derivative statistics~\citep[NGBoost,][]{duan_ngboost_2020}. Therefore, we aim to find a method that efficiently generates high-quality probabilistic forecasts with a single model using GBMs.

We address the challenge of large-scale probabilistic forecasting by proposing a novel, simple probabilistic forecasting method that leverages the popular \acl{GB} paradigm to provide accurate probabilistic forecasts in large-scale data settings (Section~\ref{sec:PGBM}). We demonstrate that our approach achieves state-of-the-art point performance as well as probabilistic performance in forecasting tasks while only training a single ensemble of Gradient Boosted Decision Trees (GBDT). Our proposed method, \acfi{PGBM}, consists of two steps: 
\begin{enumerate*}
\item We treat the leaf weights in each tree as random variables that we approximate during training via the sample mean and sample variance of the samples in each leaf, and 
\item We obtain an accurate estimate of the conditional mean and variance of our target for each sample by sequentially adding these random variables for each new tree. 
\end{enumerate*}
After training, we obtain a learned mean and variance for each sample which can be used during prediction. Based on the learned mean and variance, we can specify a distribution from which to obtain our probabilistic forecast after training. Our \ac{PGBM} is simple as its learning procedure is comparable to standard gradient boosting such as LightGBM \citep{ke_lightgbm_2017} or xgboost \citep{chen_xgboost_2016} while it requires only few additional computation steps during training and prediction. However, contrary to these existing methods, our method only requires training a single ensemble of decision trees to obtain a model capable of providing probabilistic predictions.

We empirically demonstrate that \ac{PGBM} offers state-of-the-art point and probabilistic regression performance on 11 datasets of various sizes (Section~\ref{sec:benchmark}). Therefore, PGBM provides the advantages of existing state-of-the-art point prediction gradient boosting packages such as LightGBM or xgboost, as well as the advantage of a state-of-the-art probabilistic prediction package such as NGBoost. In addition, we show how to optimise \ac{PGBM}'s probabilistic estimates \textit{after} training by varying a single hyperparameter and choosing different sets of posterior distributions. This offers the benefit of training a model only once, and optimizing for probabilistic performance thereafter. Neither existing standard gradient boosting packages or probabilistic packages such as NGBoost offer this.
Furthermore, we demonstrate that our GPU-based implementation of \ac{PGBM} trains an order of magnitude faster than existing state-of-the-art probabilistic gradient boosting methods on large datasets. Finally, we showcase the use of \ac{PGBM} on the problem of probabilistic hierarchical time series forecasting, demonstrating that our implementation enables the optimization of complex differentiable loss functions without manually specifying an analytical gradient and hessian (in contrast to existing gradient boosting packages that rely on a priori specification of an analytical gradient and hessian).

In summary, the contributions of this paper are the following:
\begin{itemize}[leftmargin=*]
	\item We introduce \ac{PGBM}, a gradient boosting framework for probabilistic regression problems (Section~\ref{sec:PGBM});
	\item We demonstrate state-of-the-art point performance and probabilistic performance of \ac{PGBM} on a set of regression benchmarks (Section~\ref{sec:benchmark}); 
	\item We show that \ac{PGBM}'s probabilistic performance can be optimized \textit{after} training the model, which allows practitioners to choose different posterior distributions without needing to retrain the model (Section~\ref{sec:benchmark});
	\item Our implementation of \ac{PGBM} trains up to several orders of magnitude faster on larger datasets than competing methods (Section~\ref{sec:benchmark}), and our implementation allows the use of complex differentiable loss functions, where we observed up to 10\% improvement in point forecasting performance and up to 300\% improvement in probabilistic forecasting performance (Section~\ref{sec:hierarchicaltimeseries}).
\end{itemize}


\section{Background}
Gradient boosting optimizes a loss function by iteratively adding a set of weak learners into an ensemble \cite{friedman_greedy_2001}. Each new weak learner is added sequentially, such that this new learner reduces the aggregate error from the existing ensemble of weak learners. Typically the weak learners are decision trees due to their strong empirical performance; and we also choose them as base learners in this work following the formalization of gradient boosting with decision trees due to \citet{chen_xgboost_2016}. In gradient boosting, at each iteration \(k\), we seek to construct a decision tree \(f^{(k)}(\vec{x}_i)\) such that the update equation for our estimate for sample \(i\) reads:
\begin{align} \label{eq:estimate}
\hat{y}^{(k)}_i &= \hat{y}^{(k - 1)}_i - \alpha f^{(k)}(\vec{x}_i),
\end{align}
in which \(\alpha\) denotes the learning rate, typically chosen to be less than \(1\), such that only a tiny portion of each new base learner is added to the overall estimate at each iteration. We will derive a different set of update equations for PGBM in Section~\ref{sec:updateequations}. 
To construct the decision tree \(f^{(k)}\), we greedily split our training data based on its input features \(\vec{x}\) into left \((I_L)\) and right \((I_R)\) nodes by maximizing the following \textit{gain}:
\begin{equation} 
\label{eq:splitdecision}
G = \frac{1}{2} \left[ \frac{(\sum_{i \in I_L} g_i)^2}{\sum_{i \in I_L}h_i + \lambda} + \frac{(\sum_{i \in I_R} g_i)^2}{\sum_{i \in I_R}h_i + \lambda} - \frac{(\sum_{i \in I} g_i)^2}{\sum_{i \in I}h_i + \lambda} \right],
\end{equation}
where \(\lambda\) is a regularization parameter, \(I = I_L \cup I_R\), and \(g_i, h_i\) are the gradient and hessian, respectively, with respect to \(\hat{y}_i^{(k-1)}\) of some differentiable loss function that we aim to minimize, for example the mean-squared error loss in case of a regression problem.\footnote{Some researchers refer to this method as \textit{Newton boosting} rather than \textit{gradient boosting} \cite{sigrist_gradient_2020}, as it employs a second-order derivative.}\textsuperscript{,}\footnote{Unlike \cite{chen_xgboost_2016}, we drop the regularization term \(\gamma T\). We have no need for this regularization parameter, as the number of leaves \(T\) is a hyperparameter that needs to be specified in our method. Note also that the parameter \(\gamma\) is in fact by default set to zero in the xgboost implementation of \cite{chen_xgboost_2016}.} When constructing the decision tree, Eq. \eqref{eq:splitdecision} is evaluated at each node to find the best possible split gain \(G^*\) among all features in the input \(\vec{x}\), and typically a split is made if the gain exceeds a certain threshold. If no split is made, the node becomes a \textit{leaf} and the corresponding leaf weight \(w_j\) follows from:
\begin{equation} 
\label{eq:leafweight}
w_j = - \frac{\sum_{i \in I_j} g_i}{\sum_{i \in I_j}h_i + \lambda},
\end{equation}
in which \(j \in \{0, 1, \dots, T\}\), with \(T\) the total number of leaves in our tree. We typically stop learning the tree if some pre-defined criterion is met, for example if no more splits with a positive split gain according to Eq. \eqref{eq:splitdecision} can be made or the tree reaches a pre-defined fixed number of leaves. After training the tree, the output \(f^{(k)}(\vec{x}_i)\) for a particular sample is then simply the leaf weight \(w_j\), or:
\begin{align} \label{eq:updateestimate}
\hat{y}^{(k)}_i = \hat{y}^{(k - 1)}_i - \alpha w_j.
\end{align}


\section{Probabilistic Gradient Boosting Machines} \label{sec:PGBM}
We discuss our method \textit{Probabilistic Gradient Boosting Machines} (PGBM). We introduce our problem setting, the core components of PGBM, and end with an analysis and discussion of PGBM.

\subsection{Probabilistic Forecasting} \label{sec:probforecasting}
In this work, we are interested in the problem of probabilistic forecasting. More generally, we are interested in the problem of probabilistic regression, in which one aims to estimate a conditional probability distribution \(P(y \vert \vec{x})\) of some target scalar variable \(y\) based on a set of inputs \(\vec{x}\). In the case of probabilistic forecasting, \(\vec{x}\) commonly includes lagged target variables as well as additional covariates. We are interested in finding a model \(f(\vec{x})\) that provides us with an estimate of the mean \(\mu\) and variance \(\sigma^2\) of a target distribution such that we can obtain a sample of an estimate \(\hat{y}\) by sampling from a specified distribution \(D\) after training:
\begin{align}
(\mu_{\hat{y}}, \sigma^2_{\hat{y}}) &= f(\vec{x}) \\
\hat{y} &\sim D(\mu_{\hat{y}}, \sigma^2_{\hat{y}}).
\label{eq:distributionsampling}
\end{align}
We construct our model \(f(\vec{x})\) using gradient boosting. In order to find the estimate for the mean and variance of our target distribution, we next derive formulas for stochastic leaf weights (Eq. \eqref{eq:leafweight}) and new update equations (Eq. \eqref{eq:updateestimate}). 

\subsection{Stochastic Leaf Weights}
By creating stochastic leaf weights, we are able to learn a mean and variance of each leaf weight in each tree, thus enabling us to learn a mean and variance for each sample in our dataset. We assume that the gradient and hessian of our loss function are random variables with a mean \((\mu_g, \mu_h)\) and finite variance \((\sigma^2_g, \sigma^2_h)\) and covariance \(\sigma^2_{gh}\), which we approximate separately in each tree for each instance set \(I_j\) using the sample mean, sample variance and sample covariance for the \(n_j\) samples in an instance set \(I_j\):
\begin{align} \label{eq:samplestatistics}
\mu_{g} &\approx \overline{g}_j = \frac{1}{n_j} \sum_{i \in I_j} g_i \\
\mu_{h} &\approx \overline{h}_j = \frac{1}{n_j} \sum_{i \in I_j} h_i \\
\sigma_g^2 &\approx \overline{\sigma}^2_{g_j} = \frac{1}{n_j - 1}\sum_{i \in I_j}(g_i - \overline{g})^2 \\
\sigma_h^2 &\approx \overline{\sigma}^2_{h_j} = \frac{1}{n_j - 1}\sum_{i \in I_j}(h_i - \overline{h})^2 \\
\sigma_{gh}^2 &\approx \overline{\sigma}^2_{{gh}_j} = \frac{1}{n_j - 1}\sum_{i \in I_j}(g_i - \overline{g})(h_i - \overline{h}).
\end{align}
Note that the sample variance and covariance require the Bessel correction \(n_j - 1\) in order to obtain an unbiased estimate of the true variance and covariance. Moreover, the Central Limit Theorem dictates that we obtain our true mean and variance if \(n_j \to \infty\). However, for tiny datasets, \(n_j\) is typically a small number as each leaf may only contain a few samples and thereby using sample statistics might be inappropriate. In the Experiments section (Section~\ref{sec:experiments}), we will demonstrate that we are still able to provide accurate probabilistic estimates even in such cases. Next, we write Eq. \eqref{eq:leafweight} in terms of the sample mean of the gradient and hessian:
\begin{align}
w_j = - \frac{\frac{1}{n_j} \sum_{i \in I_j} g_i}{\frac{1}{n_j} \sum_{i \in I_j}h_i + \frac{1}{n_j} \lambda} = - \frac{\overline{g}_j}{\overline{h}_j + \overline{\lambda}_j}.
\end{align}
Now, we can model the expectation and variance of the leaf weight \(w_j\) using the sample statistics as follows (dropping the subscript \(j\) for readability):
\begin{align}
\mu_j = E\left[\frac{\overline{g}}{(\overline{h} + \overline{\lambda})}\right]  &\approx \frac{\overline{g}}{(\overline{h} + \overline{\lambda})} - \frac{\overline{\sigma}^2_{gh}}{(\overline{h}  + \overline{\lambda})^2} +  \frac{\overline{g} \overline{\sigma}^2_h}{(\overline{h} +  \overline{\lambda})^3} \label{eq:expectation} \\
\sigma^2_j = V\left[\frac{\overline{g}}{(\overline{h} + \overline{\lambda})}\right] &\approx  \frac{\overline{\sigma}^2_g}{(\overline{h} + \overline{\lambda})^2} + \frac{\overline{g}^2 \overline{\sigma}_h^2}{(\overline{h} + \overline{\lambda})^{4}} - 2 \frac{\overline{g}\overline{\sigma}^2_{gh}}{(\overline{h} + \overline{\lambda})^{3}}. 
\label{eq:variance}
\end{align}
We refer the reader to the Supplemental Materials~\ref{app:stweights} for the derivation of Eq.~\eqref{eq:expectation}--\eqref{eq:variance}. Note that for most common loss functions such as the mean-squared error, the two final terms of Eq.~\eqref{eq:expectation}--\eqref{eq:variance} are zero as the Hessian \(h\) has no variation. When training a decision tree \(f^{(k)}\), we store the obtained expectation and variance of each leaf of each tree and use these results to obtain our final estimate for the mean and variance using the update equations described in the next subsection. 

\subsection{Update Equations} 
\label{sec:updateequations}
Apart from the stochastic leaf weights, we require new update equations (Eq.~\eqref{eq:updateestimate}) in order to update the estimate for our mean and variance when adding a new tree at each iteration. These new update equations allow us to aggregate the stochastic weights over all the trees. For these equations, we make use of the following rules for the mean \(\mu\) and variance \(\sigma^2\) of some random variables \((A, B)\) and a constant \(c\):	
\begin{align*}
\mu_{(A - cB)} &= \mu_A - c \cdot \mu_B \\
\sigma^2_{cB} &= c^2 \sigma^2_{B} \\
\sigma^2_{(A - cB)} &=  \sigma^2_{A} + c^2 \sigma^2_{B} - 2 c \cdot \sigma^2_{(A,B)} \\
\sigma^2_{(A,B)} &= \rho_{(A,B)} \sigma_A \sigma_B,
\end{align*}
in which \(\rho\) denotes Pearson's correlation coefficient between the variables \((A, B)\). Using these rules, we can modify Eq.~\eqref{eq:updateestimate} to arrive at the formulas for the expectation \(E\) and variance \(V\) of our estimate~\(\hat{y}^{(k)}_i\):
\begin{align} \label{eq:updateequations}
\mu_{\hat{y}_i^{(k)}} &= E\left[\hat{y}^{(k)}_i\right] = \mu_{\hat{y}_i^{(k-1)}} - \alpha \cdot \mu_{j^{(k)}} \\
\sigma^2_{\hat{y}_i^{(k)}} &= V\left[\hat{y}^{(k)}_i\right] = \sigma^2_{\hat{y}_i^{(k-1)}} + \alpha^2 \sigma^{2}_{j^{(k)}} - 2 \alpha \rho \sigma_{\hat{y}_i^{(k-1)}} \sigma_{j^{(k)}}.
\label{eq:updateequationvariance}
\end{align}
where the hyperparameter \(\rho\) denotes the correlation coefficient between trees \(k\) and \(k-1\). We provide further discussion around \(\rho\) in Section~\ref{sec:discussion}. Finally, the learned expectation and variance can be used for creating probabilistic predictions of new samples after training our model by sampling from a distribution parameterized by these learned quantities:
\begin{align}
\hat{y}^{(k)}_i &\sim D\left( \mu_{\hat{y}^{(k)}_i}, \sigma^2_{\hat{y}^{(k)}_i} \right),
\end{align}
 We are now ready to fully present our method PGBM.

\subsection{PGBM}
\header{Algorithm} We provide a succint overview of the procedures for training and prediction with Probabilistic Gradient Boosting Machines (PGBM) in Algorithms~\ref{alg:pgbm_training}\&~\ref{alg:pgbm_prediction}.

\header{\textit{Training}~(\textit{Algorithm~\ref{alg:pgbm_training}})} In PGBM, gradient boosting is performed comparable to LightGBM~\cite{ke_lightgbm_2017} or xgboost \cite{chen_xgboost_2016}, and PGBM employs global equal density histogram binning to bin continuous features into discrete bins in order to reduce the computational effort required to find the optimal split decision (Line~\ref{algline:binning}). At the start of training, we initialize the estimate \(\hat{\vec{y}}\), typically with the mean of the training set in a regression setting (Line~\ref{algline:initial}). Then, gradient boosting is performed for a fixed number of iterations by first computing the gradient and hessian (Lines~\ref{algline:computegradient}--\ref{algline:computehessian}) of the training set and optionally choosing a subsample of the dataset (commonly referred to as bootstrapping, Line~\ref{algline:bootstrapping}) on which to build the decision tree. The decision tree is then constructed up to a fixed number of leaves (Line~\ref{algline:fixedleaves}) by first selecting the samples in the current node (Line~\ref{algline:currentnode}), second by finding the best split for this node (Line~\ref{algline:findbestsplit}), and third by splitting the current node or creating stochastic leaf weights if no split can be made (Line~\ref{algline:splitdecision}), for example, when the split does not result in a positive gain according to Eq.~\eqref{eq:splitdecision}. After the tree construction has finished, predictions for the entire training set are generated (Line~\ref{algline:predicttrainingset}) and the overall estimate is updated (Line~\ref{algline:updateestimate}) and the process repeats for the next iteration. 

Note that the learned variance is only used to create the probabilistic estimate in the prediction algorithm; it can also serve as a validation criterion during training (for example, by performing a prediction step on a validation set and deciding based on some probabilistic metric whether to continue training or not).

\header{\textit{Prediction~(Algorithm~\ref{alg:pgbm_prediction}})} During prediction, we initialize the estimate using the stored initial estimate of the training set (Line~\ref{algline:initialprediction}). Then, we make predictions on the dataset by iterating over all the trees (Line~\ref{algline:iteratepredictions}) using our new update equations (Line~\ref{algline:updatepredictions}). Finally, we obtain our probabilistic estimate by sampling from a distribution parameterized by our learned mean and variance (Line~\ref{algline:predictdistr}).

\begin{algorithm}
\caption{PGBM training algorithm}
\label{alg:pgbm_training}
\begin{flushleft}
\textbf{Input}: Input dataset \(\vec{X} \in \mathbb{R}^{n \times f}\) with \(n\) samples and \(f\) features, target output \(\vec{y} \in \mathbb{R}^{n}\), differentiable loss function \(l(\vec{y}, \vec{\hat{y}})\) and model \texttt{hyperparameters}. \\
\textbf{Output}: 
\end{flushleft}
\begin{algorithmic}[1]
\STATE Bin features such that for each feature \( \vert \vec{x} \vert \le \texttt{max\_bins} \) \label{algline:binning}
\STATE Set initial estimate \(\hat{\vec{y}}\), e.g. to mean \(\overline{\vec{y}}\) of target output \label{algline:initial}
\FOR { \(k = 1\)  \TO num\_iterations }
	\STATE Compute gradient \(\vec{g}^{(k)} = \nabla_{\vec{\hat{y}}^{(k)}} l(\vec{y}, \vec{\hat{y}}) \) \label{algline:computegradient}
	\STATE Compute hessian \(\vec{h}^{(k)} = \nabla^2_{\vec{\hat{y}}^{(k)}} l(\vec{y}, \vec{\hat{y}}) \) \label{algline:computehessian}
	\STATE Select subsample of input dataset as instance set \(I_1\) \label{algline:bootstrapping}
	\FOR { \(j = 1\) \TO max\_leaves} \label{algline:fixedleaves}
		\STATE Select instance set \(I_j\) of \( \vec{X}, \vec{g}^{(k)}, \vec{h}^{(k)} \) \label{algline:currentnode}
		\STATE Find best split for all (features, bins) (Eq.~\eqref{eq:splitdecision}) \label{algline:findbestsplit}
		\STATE Create split if split criteria are met else create stochastic leaf weight (Eq. \eqref{eq:samplestatistics}--\eqref{eq:variance}) \label{algline:splitdecision}
	\ENDFOR
	\STATE Predict \(\vec{X}\) to obtain \(\mu^{(k)}_{\vec{\hat{y}}}\) (Eq.~\eqref{eq:updateequations}) \label{algline:predicttrainingset}
	\STATE Update estimate \(\hat{\vec{y}} =\mu^{(k)}_{\vec{\hat{y}}}\)  \label{algline:updateestimate}
\ENDFOR
\end{algorithmic}
\end{algorithm}
\vspace{-5mm}

\begin{algorithm}
\caption{PGBM prediction algorithm}
\label{alg:pgbm_prediction}
\begin{flushleft}
\textbf{Input}: Input dataset \(\vec{X} \in \mathbb{R}^{n \times f}\) with \(n\) samples and \(f\) features, target distribution \(D\) and model \texttt{hyperparameters}. \\
\textbf{Output}: 
\end{flushleft}
\begin{algorithmic}[1]
\STATE Set initial estimate \(\hat{\vec{y}}\) to mean \(\overline{\vec{y}}\) of training dataset \label{algline:initialprediction}
\FOR { \(k = 1\)  \TO num\_iterations} \label{algline:iteratepredictions}
	\STATE Predict \(\vec{X}\) to obtain \((\mu^{(k)}_{\vec{\hat{y}}}, \sigma^{2(k)}_{\vec{\hat{y}}} )\) (Eq.~\eqref{eq:updateequations}--\eqref{eq:updateequationvariance}) \label{algline:updatepredictions}
\ENDFOR
\STATE Draw \texttt{n\_samples} \(\vec{\hat{y}} \sim D\left( \mu_{\vec{\hat{y}}^{(k)}}, \sigma^2_{\vec{\hat{y}}^{(k)}} \right)\) \label{algline:predictdistr}
\end{algorithmic}
\end{algorithm}

\header{Implementation} We implement PGBM in PyTorch \cite{paszke_pytorch_2019} and offer it as a Python package.\footnote{\url{https://github.com/elephaint/pgbm}} PyTorch offers (multi-)GPU acceleration by default, which allows us to scale PGBM to problems involving a large number of samples (we trained on datasets of over 10M samples) as we can distribute training across multiple GPUs. More importantly, our implementation allows the use of the automated differentiation engine of PyTorch, such that we can employ arbitrary complex differentiable loss functions without requiring an analytical gradient and hessian. This is in stark contrast to existing popular packages such as LightGBM~\cite{ke_lightgbm_2017} or xgboost~\cite{chen_xgboost_2016}, where custom loss functions require the manual derivation of an analytical gradient and hessian. We provide an example of this benefit in Section~\ref{sec:hierarchicaltimeseries}. Note that PGBM can be made compatible with existing gradient boosting packages relatively easily too, as it only requires storing one additional sample statistic (the variance), changing the update equations according to Section~\ref{sec:updateequations} and choosing a distribution \(D\) after training to sample from. We intend to provide an implementation of PGBM within existing popular gradient boosting packages in the future. 

Furthermore, we implemented a custom CUDA kernel that integrates with PyTorch to calculate the optimal splitting decision (Eq.~\eqref{eq:splitdecision}), the most compute intensive part of PGBM. Our kernel leverages the parallel processing power of modern CUDA-capable GPUs, by parallelizing the split decision across the sample and feature dimension using parallel reductions. We demonstrate the effectiveness of our GPU training in Section~\ref{sec:benchmark}.

\subsection{Analysis \& Discussion} \label{sec:discussion}

\header{Computational complexity} Even though two parameters -- a mean and variance -- are learned in PGBM, the trees are constructed comparable to standard gradient boosting such as in \cite{ke_lightgbm_2017, chen_xgboost_2016}. Therefore, the additional cost of our second parameter is negligible as only the sample statistics need to be calculated in the leaves. In contrast to NGBoost~\cite{duan_ngboost_2020}, PGBM also does not require calculation of a natural gradient, which involves the inversion of many small matrices. PGBM's runtime generally scales with the number of samples, the number of features and the number of bins used to bin the features, in accordance with existing GBM packages.

\header{Higher-order moments and leaf sample quantiles} We only consider the first two moments of a distribution (i.e., the mean and variance) to derive our stochastic leaf weights, which limits the output distribution \(D\) to distributions parameterized using location and scale parameters (i.e., our learned mean and variance). This is a limitation compared to, e.g., NGBoost \cite{duan_ngboost_2020}. We considered calculating higher order sample statistics such as the sample skewness (third moment) and sample kurtosis (fourth moment), however the disadvantage is that 
\begin{enumerate*}[leftmargin=*,label=(\roman*)]
\item there is no unbiased sample statistic for those measures, 
\item deriving approximations of the form of Eq.~\eqref{eq:expectation}--\eqref{eq:variance} becomes exceedingly complex, and 
\item higher-order sample statistics require more samples in order for the sample statistic to provide a reasonable estimate of the true statistic. 
\end{enumerate*}
Moreover, as we learn separate sample statistics for each leaf in each tree, we are still able to model complex distributions over the entire dataset using distributions parameterized only by location and scale parameters. 

Finally, one could also store the sample quantile information of each leaf and draw samples according to the stored quantile information. This would remove the need for specifying a particular distribution. While this seems an attractive option, calculating sample quantile information for each leaf is computationally difficult as it requires an expensive sorting operation, and storing a sufficient number of sample quantiles to reap the full benefits of this method requires storing at least 2--3x the number of leaf weights. In short, there is no real need to use higher order moments or leaf sample quantiles to provide accurate probabilistic estimates as we demonstrate in Section~\ref{sec:benchmark}.

\header{Output sampling} PGBM allows one to sample from different output distributions \textit{after} training, which allows practitioners to train their model by minimizing some point metric (e.g., RMSE) and after training try different distributions for optimizing the probabilistic forecast based on some validation metric. The key benefit is that this allows PGBM to achieve state-of-the-art point forecasting performance \textit{as well as} accurate probabilistic forecasting performance \textit{using the same model}. We will demonstrate this in Section~\ref{sec:benchmark}. Note that practitioners can of course also directly optimize the probabilistic forecast by using a loss function that optimizes the probabilistic forecast. 

\header{Split decisions and tree dependence} In PGBM, split decisions in the tree are not recomputed based on the stochasticity of the leaf weights, even though it could be argued that this would be appropriate when sampling from the trees. We intentionally avoid this as it is computationally expensive to recompute split decisions after training when sampling from the learned distribution. Secondly, by sequentially adding the mean and variance of each tree we omit the explicit covariance between tree \(k\) and trees \(k - 2, k - 3, \dots\), and only model the covariance between subsequent trees. We implicitly model both these effects using a single constant correlation hyperparameter \(\rho\) (Eq.~\eqref{eq:updateequationvariance}), for which we provide a more detailed analysis in Section~\ref{sec:benchmark}.

\header{Hessian distribution} The distribution of the Hessian \(h\) should have a support of \([0, \infty)\) to avoid division by zero in Eq. \eqref{eq:expectation}--\eqref{eq:variance}, or equivalently, we require the sum of the hessians (plus regularization constant \(\lambda\)) of all samples in an instance set \(I_j\) of a leaf to be positive. For common convex loss functions such as the mean-squared error this is not an issue, however for non-convex loss functions this might pose a problem in rare cases where all hessians in an instance set add up to zero. In those cases, numerical issues can usually be avoided by requiring a decent (e.g., $>10$) minimum number of samples in each leaf in a tree -- this can be set as a hyperparameter in PGBM.


\section{Experiments}
\label{sec:experiments}
In this section, we first demonstrate how PGBM can provide accurate point and probabilistic predictions on a set of common regression benchmark datasets from the UCI Machine Learning Repository (Section~\ref{sec:benchmark}). We show how PGBM allows practitioners to optimize their probabilistic estimate after training, thereby removing the need to retrain a model under different posterior distribution assumptions. Next, we demonstrate the efficiency of our implementation of PGBM compared to existing gradient boosting methods. Finally, we demonstrate PGBM on the problem of forecasting for hierarchical time series, which requires optimizing a complex loss function for which deriving an analytical gradient is too complex (Section~\ref{sec:hierarchicaltimeseries}). 

To facilitate reproducibility of the results in this paper, our experiments are run on open data, and our experimentation code is available online.\footnote{Repository at \url{https://github.com/elephaint/pgbm}}

\subsection{UCI regression benchmarks} \label{sec:benchmark}

\header{Task} We perform probabilistic regression on a set of regression datasets. Our goal is to obtain the lowest probabilistic prediction score as well as the lowest point performance score.

\header{Evaluation} We evaluate the probabilistic performance of each method using the Continuously Ranked Probability Score (CRPS), which is a measure of discrepancy between the empirical cumulative distribution function of an observation \(y\) and the cumulative distribution \(F\) of a forecast \(\hat{y}\) \cite{zamo_estimation_2018}:
\begin{equation}
C = \int [F(\hat{y}) - \mathbbm{1}(\hat{y} \ge y)]^2 d\hat{y},
\end{equation}
in which \(\mathbbm{1}\) denotes the indicator function. We compute the empirical CRPS based on 1,000 sample predictions generated by the trained models on the test set. We evaluate point performance using Root Mean Squared Error (RMSE): 
\begin{equation}
RMSE = \sqrt{\frac{1}{n} \sum_i^n (y_i - \hat{y}_i)^2}.
\end{equation}
We present these metrics relative to the median of PGBM over all the folds tested for a dataset, and we refer the reader to Table~\ref{tab:ucibenchmark} of Supplemental Materials~\ref{app:reproducibility} for further details.

\header{Protocol} We follow the same protocol as \citet{duan_ngboost_2020}, and create 20 random folds for each dataset except for \texttt{msd} for which we only create one. For each of these folds, we keep 10\% of the samples as test set. The remaining 90\% is first split into an 80/20 validation/training set to find the number of training iterations that results in the lowest validation score. After validation, the full 90\% training set is trained using the number of iterations found in the validation step. As output distribution for the probabilistic prediction we use a Normal distribution, similar to \citet{duan_ngboost_2020}.

\header{Baseline models} For probabilistic performance, we compare against NGBoost \cite{duan_ngboost_2020}, which has recently been shown to outperform other comparable methods on the current set of benchmarks. We use the same settings for NGBoost as in \cite{duan_ngboost_2020}. For point performance, we also compare to LightGBM \cite{ke_lightgbm_2017}, one of the most popular and best-performing gradient boosting packages available. We configure LightGBM to have the same settings as PGBM. 

\header{PGBM} For all datasets, we use the same hyperparameters for PGBM, except that we use a bagging fraction of 0.1 for \texttt{MSD} in correspondence with \citet{duan_ngboost_2020}. Our training objective in PGBM is to minimize the mean-squared error (MSE). We refer the reader to Table~\ref{tab:exphyperparam} of Supplemental Materials~\ref{app:reproducibility} for an overview of key hyperparameters of each method.

\begin{figure*}
\centering
\captionsetup{font=small}
\begin{subfigure}{\columnwidth}
  \centering
  \includegraphics[width=\columnwidth]{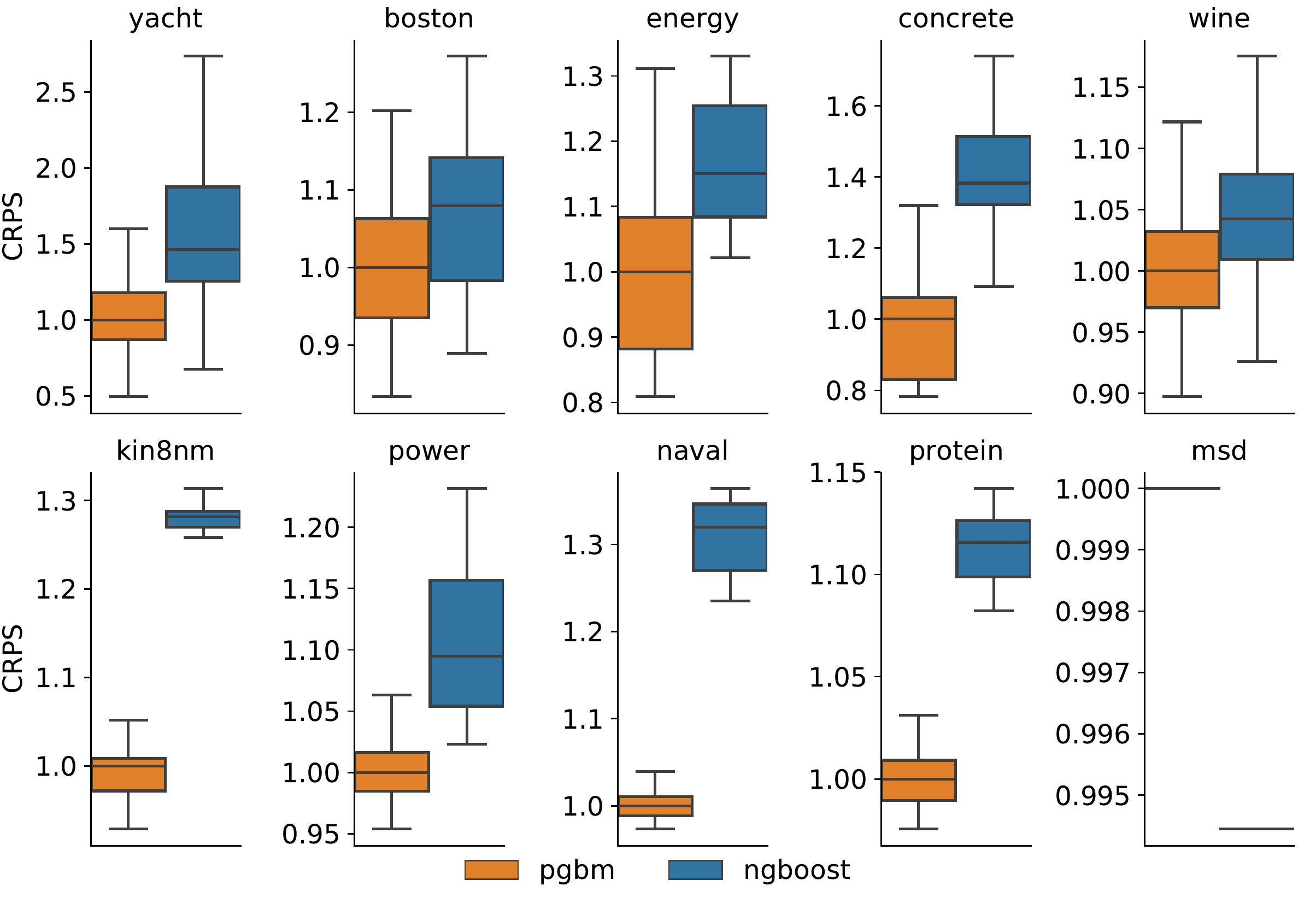}
  \caption{Probabilistic performance (CRPS)}
  \label{fig:crps}
\end{subfigure}
\begin{subfigure}{\columnwidth}
  \centering
  \includegraphics[width=\columnwidth]{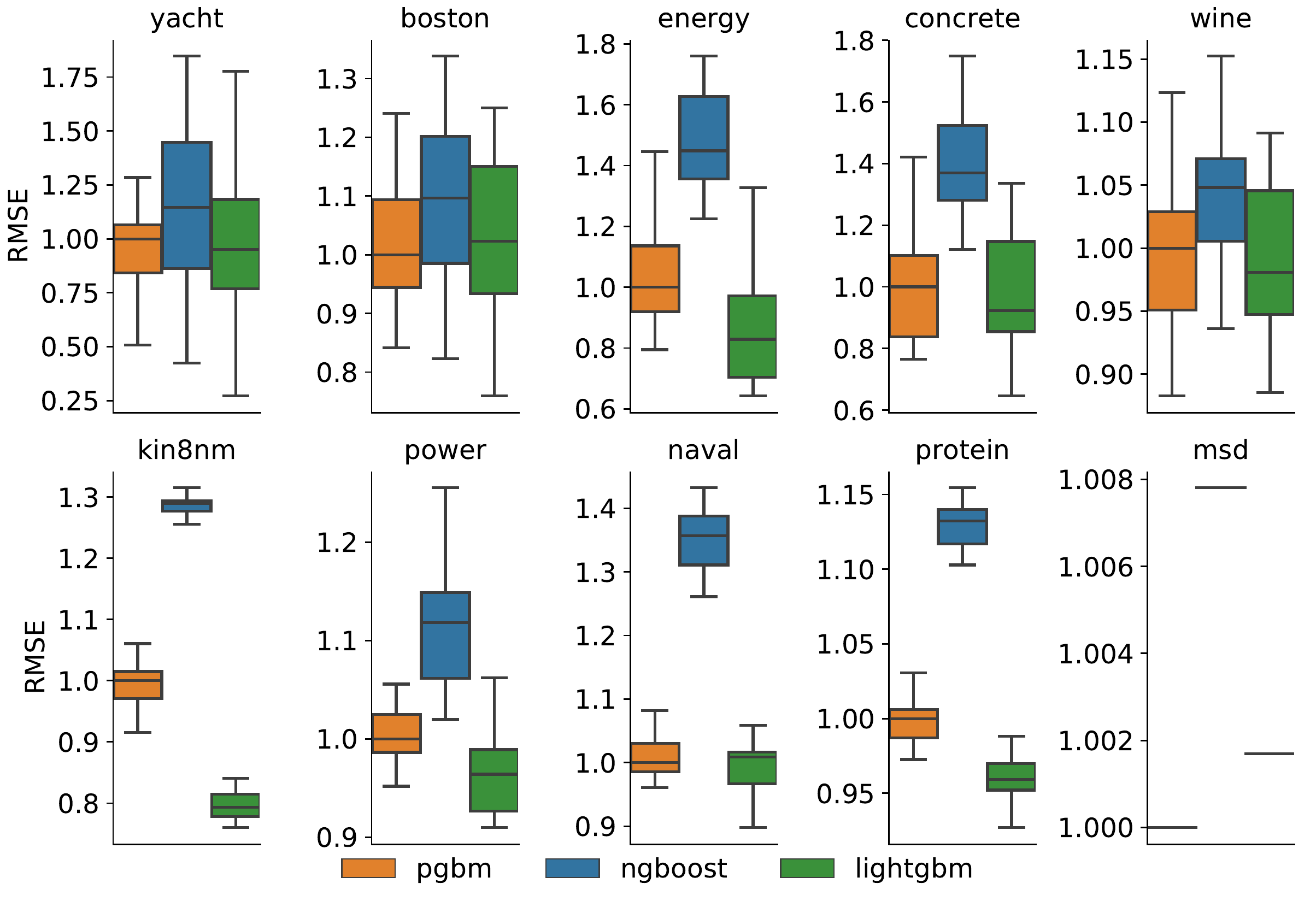}
  \caption{Point performance (RMSE)}
  \label{fig:rmse}
\end{subfigure}
\caption{Results for probabilistic (CRPS, left) and point (RMSE, right) performance for each dataset for each method, with the smallest dataset \texttt{yacht} in the top left corner and the largest dataset \texttt{msd} in the lower right corner. Lower is better, and results have been indexed against the median test score of PGBM. PGBM outperforms NGBoost both in probabilistic and point performance.}
\label{fig:exp1results}
\end{figure*}

\header{Results} We provide the results of our first experiment in Figure~\ref{fig:exp1results} and observe the following: 
\begin{itemize}[leftmargin=*]
	\item On probabilistic performance, PGBM outperforms NGBoost on average by approximately 15\% as demonstrated by the relatively lower CRPS across all but one dataset (\texttt{msd}, where the difference is tiny). This is remarkable, as the training objective in PGBM is to minimize the MSE, rather than optimize the probabilistic forecast as does NGBoost.
	\item PGBM outperforms NGBoost on all datasets on point performance, and on average by almost 20\%, which is in line with expectation as we explicitly set out to optimize the MSE as training objective in PGBM in this experiment. However, as becomes clear from this result, PGBM does not have to sacrifice point performance in order to provide state-of-the-art probabilistic estimates nonetheless. Compared to LightGBM, PGBM performs slightly worse on average (approx. 3\%) on point performance. We suspect this is due to implementation specifics. 
\end{itemize}
The main takeaway from this experiment is that even though we only optimized for a point metric (MSE) in PGBM, we were still able to achieve similar probabilistic performance compared to a method that explicitly optimizes for probabilistic performance. 

\begin{figure}
\centering
\captionsetup{font=small}
\begin{subfigure}{0.49\columnwidth}
  \centering
  \includegraphics[width=\columnwidth]{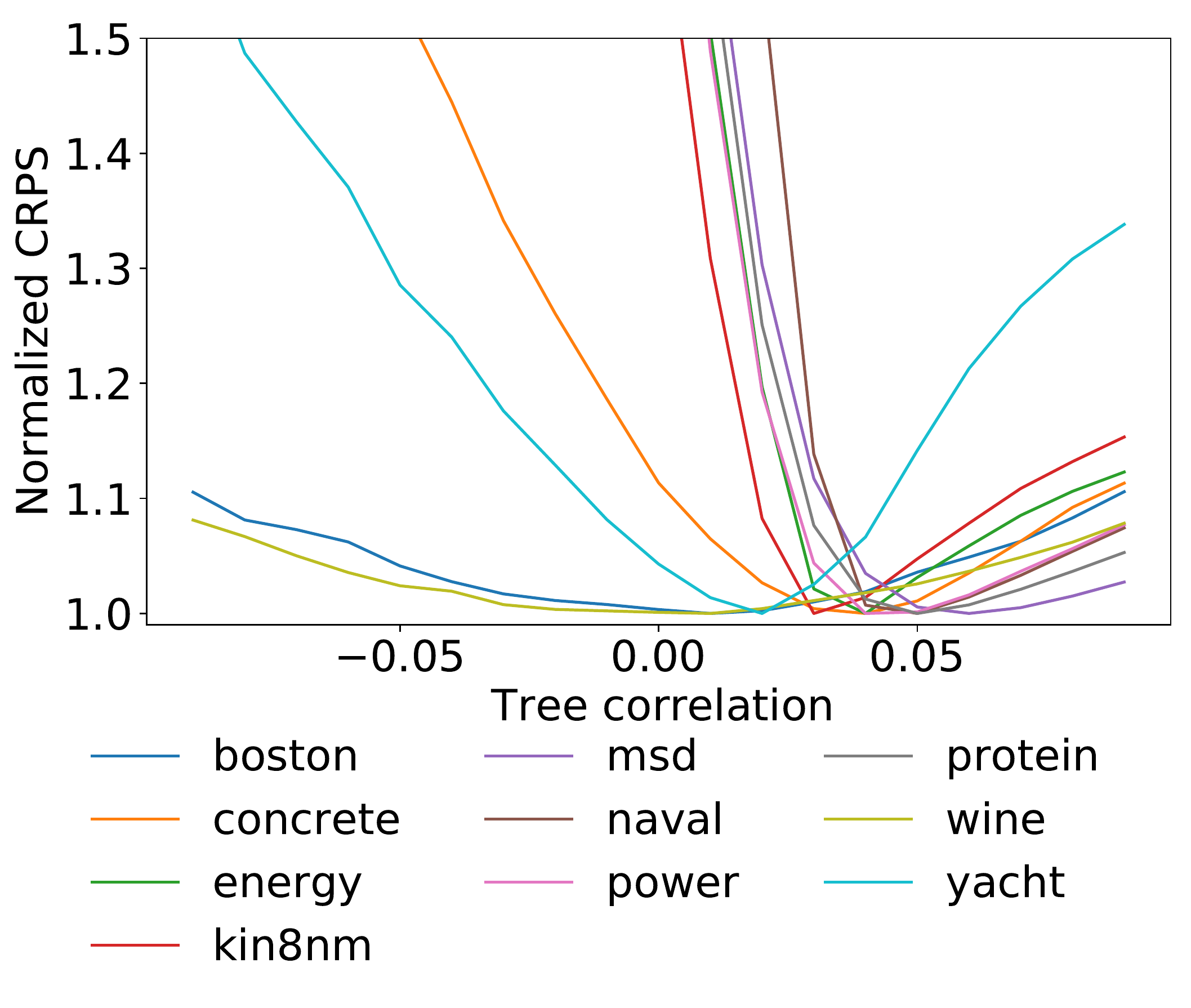}
  \caption{}
  \label{fig:figcrpsvscorr}
\end{subfigure}
\begin{subfigure}{0.49\columnwidth}
  \centering
  \includegraphics[width=\columnwidth]{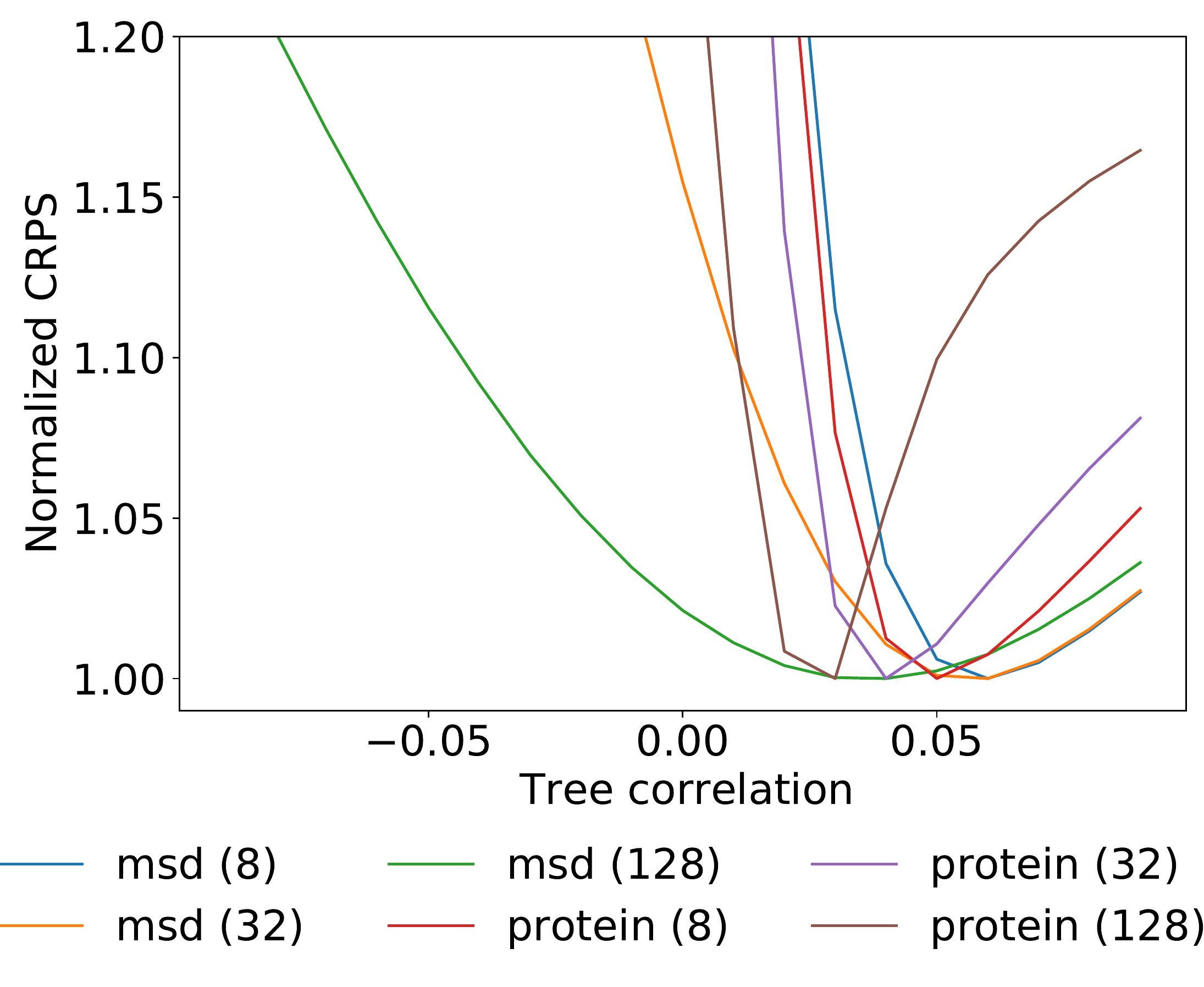}
  \caption{}
  \label{fig:figcrpsvscorr_maxleaves}
\end{subfigure}
\caption{Normalized CRPS on the validation set for different settings of tree correlation hyperparameter \(\rho\), (a) for all datasets and (b) for \texttt{protein} and \texttt{msd} when trained using a maximum number of leaves per tree of \(\{8, 32, 128\}\).}
\end{figure}

\header{Analysis: correlation hyperparameter} We perform a brief analysis of the correlation hyperparameter \(\rho\) (Eq.~\eqref{eq:updateequationvariance}). This hyperparameter controls the dependence between variance estimates of subsequent trees in PGBM, and is critical for probabilistic performance. Figure~\ref{fig:figcrpsvscorr} shows the CRPS evaluated on the validation set at different settings for \(\rho\) for each dataset for a single fold. We normalized the CRPS scores on the lowest CRPS for each dataset. Across all datasets, the CRPS seems to follow a parabolic shape and consequently there seems to be an optimal choice for \(\rho\) across different datasets: a value of \(0.02\)--\(0.07\) typically seems appropriate. Empirically, we found that an initial value of \(\rho = \frac{\log_{10} n}{100}\), where \(n\) denotes the size of the training set generally works well and therefore we used that in our experiments as default value. Intuitively, the positiveness of the correlation between subsequent trees seems logical: if the leaf weight of a given tree shifts more positively (negatively) as a result of stochasticity, the residual on which the next tree will be constructed shifts in the same direction. Consequently, the leaf weights of this next tree will also shift in the same direction, thus exhibiting a positive correlation with the previous tree's leaf weights. 
Furthermore, larger datasets tend to cluster together in behavior, as can be seen from the curves for the \texttt{protein}, \texttt{msd}, \texttt{kin8nm}, \texttt{power} and \texttt{naval} datasets. It seems that for larger datasets, typically larger settings for \(\rho\) are appropriate. Our hypothesis is that for trees containing many samples per leaf, the correlation between subsequent trees is higher, as more samples in the tree's leaves will generally imply that the model has not yet (over)fit to the training set and there is likely more information left in the residuals compared to the situation where there are few samples per leaf. This would explain the behavior observed in Figure~\ref{fig:figcrpsvscorr} as we train each model with a maximum of 8 leaves per tree, resulting in more samples per leaf for larger datasets. We test this hypothesis by training the relatively larger \texttt{protein} and \texttt{msd} datasets using different settings for the maximum number of leaves, for which we show the results in Figure~\ref{fig:figcrpsvscorr_maxleaves}. Confirming our hypothesis, we indeed observe the optimal correlation parameter decreasing when we train PGBM using a higher number of maximum leaves per tree (i.e., the minimum of the parabola shifts to the left). 

\header{Analysis: posterior distribution} One of the key benefits of PGBM is that it allows us to optimize the probabilistic estimate \textit{after} training, as the choice of distribution \(D\) in Eq. \eqref{eq:distributionsampling} is independent from the training process. This offers the benefit of training to optimize a certain point metric such as RMSE, and choosing the distribution that best fits the learned mean and variance after training by validating a set of distribution choices on a validation set. To demonstrate this benefit, we repeated the experiments from our first experiment for a single fold. For each dataset, we evaluated the learned model on CRPS on the validation set for a set of common distributions and a range of tree correlations \(\rho = \{0.00, 0.01,\dots,0.09\}\). The optimal choice of distribution and tree correlation on the validation set was subsequently used for calculating the CRPS on the test set. We report the results in Table~\ref{tab:postdistribution}, where `Base case' refers to the base case scenario from our first experiment, where we chose a Normal distribution and a tree correlation hyperparameter of \(\rho = \frac{\log_{10} n}{100}\) across all datasets, and `Optimal' refers to the result on the test set when choosing the distribution and tree correlation according to the lowest CRPS on the validation set. We see that for most datasets, the minimum CRPS on the validation set is similar across choices of distribution, which implies that it is more beneficial to optimize the tree correlation rather than the choice of output distribution for these datasets. On the test set, we see improved scores compared to the base case on all but the smallest dataset, thereby showcasing the benefit of optimizing the probabilistic forecast after training. We would advice practitioners to start with a generic distribution such as the normal or Student's t(3), and optimize the probabilistic estimate after training by testing for different tree correlations and distribution choices. 

\begin{table*}[t]
\captionsetup{font=small}
\caption{Results for probabilistic (CRPS) performance for each dataset when varying the posterior distribution and tree correlations on the validation set (left) and subsequently using the optimal choice of distribution and tree correlation on the test set (right). Best results on validation and test set are in bold. We report the minimum CRPS on the validation set across all the tree correlations tested per distribution.}
\label{tab:postdistribution}
\begin{center}
\small
\begin{tabular}{l c c c c c c c c c | c c}
\toprule 
	& \multicolumn{9}{ c }{Validation set} &  \multicolumn{2}{ c }{Test set} \\
 \cmidrule(r){2-10} \cmidrule(l){11-12}
Dataset  			& Normal & Student's t(3) & Logistic & Laplace & LogNormal & Gumbel & Weibull & Poisson & NegativeBinomial & Base case & Optimal \\
\midrule
\texttt{yacht}		& \textbf{0.37}	& 0.37	& 0.37	& 0.37	& 0.44	& 0.38	& 0.37	& 0.75	& 9.7	& \textbf{0.18}	& 0.19\\
\texttt{boston}	& 1.41	& 1.4	& 1.39	& \textbf{1.39}	& 1.4	& 1.42	& 1.39	& 1.58	& 12.27	& 2.08	& \textbf{2.06} \\
\texttt{energy}	& 0.62	& 0.61	& 0.62	& \textbf{0.61}	& 0.62	& 0.62	& 0.63	& 1.25	& 16.15	& 0.64	& \textbf{0.64} \\
\texttt{concrete}	& 2.21	& 2.21	& 2.21	& \textbf{2.19}	& 2.23	& 2.26	& 2.22	& 2.38	& 3.59	& 2.69	& \textbf{2.64} \\
\texttt{wine}		& \textbf{0.32}	& 0.33	& 0.32	& 0.32	& 0.32	& 0.32	& 0.32	& 0.59	& 0.64	& 0.35	& \textbf{0.34} \\
\texttt{kin8nm}	& 0.08	& 0.08	& \textbf{0.08}	& 0.08	& 1.04	& 0.08	& 0.08	& 0.26	& 0.26	& 0.08	& \textbf{0.08} \\
\texttt{power}	& 1.86	& 1.86	& \textbf{1.86}	& 1.86	& 1.86	& 1.87	& 1.88	& 5.22	& 454.34	& 1.81	& \textbf{1.80} \\
\texttt{naval}	 	& 0.00	& 0.00 	& 0.00 	& \textbf{0.00}   & 0.02 	& 0.00 	& 0.00  & 0.22 	& 0.22 	& 0.00	& \textbf{0.00} \\
\texttt{protein}	& 2.18	& 2.18	& 2.18	& 2.18	& 45.3	& 2.17	& 2.54	& 2.15	& \textbf{2.15}	& 2.14	& \textbf{2.12} \\
\texttt{msd}		& 4.74	& 4.73	& 4.74	& 4.73	& 4.74	& 4.88	& \textbf{4.69}	& 11.16	& 1982.19	& 4.78	& \textbf{4.74} \\
\bottomrule
\end{tabular}
\end{center}
\vspace{-4mm}
\end{table*}

\header{Analysis: training time} Our implementation in PyTorch allows us to use GPU training by default, which allows us to significantly speed up training for larger datasets. We demonstrate this benefit in Table~\ref{tab:timing} where we compare training times for datasets of different size against a baseline of PGBM (we refer to Table~\ref{tab:abstiming} in the Supplemental Materials for the absolute timings). For this experiment, we also included the \texttt{higgs} dataset, which is a 10M sample UCI dataset commonly used to benchmark gradient boosting packages. For PGBM, we show results for training on GPU-only and CPU-only. NGBoost does not offer GPU training and runs on top of the default scikit-learn decision tree regressor. We ran our experiments on a 6-core machine with a nVidia RTX 2080Ti GPU. As can be seen, PGBM is up to several orders of magnitude faster than NGBoost as the dataset size increases. This demonstrates that PGBM and our implementation allow practitioners to solve probabilistic regression problems for datasets much larger than NGBoost. 

In general, for smaller datasets, training on cpu offers the best timings for the two methods. We include the relative timings to LightGBM for reference in Table~\ref{tab:timing}, which shows that PGBM even becomes competitive to LightGBM for the largest dataset (\texttt{higgs}). However, the timings for LightGBM represent the timings to train a \textit{single} LightGBM model. If one is interested in obtaining a probabilistic forecast, the timings would be multiplied by the number of quantiles required. Hence, for a fine-grained probability distribution, the timings would be 5--10x higher for LightGBM, again demonstrating the effectiveness of our implementation for probabilistic forecasting.

\begin{table}
\small
\captionsetup{font=small}
\caption{Training time for 2,000 iterations for 5 datasets of different size as a fraction of PGBM-gpu training time. Bold indicates probabilistic forecasting method with the lowest training time.}
\label{tab:timing}
\begin{center}
\begin{tabular}{l c c c | c}
\toprule 
		& \multicolumn{3}{c}{Probabilistic forecast}	&Point forecast  \\
Dataset  & \textit{PGBM-gpu} &\textit{PGBM-cpu} & \textit{NGBoost} & \textit{LightGBM} \\
\midrule
\texttt{wine} (n=1,599)  &	1.00 &	0.41 &\textbf{0.20} &0.01 \\
\texttt{naval} (n=12k) &	\textbf{1.00} &	1.62 &1.05 & 0.02 \\
\texttt{protein} (n=46k) & \textbf{1.00} &	3.09 &3.39 & 0.02 \\
\texttt{msd} (n=515k) & \textbf{1.00} &26.85 &48.83 &0.09 \\
\texttt{higgs} (n=10,5M) & \textbf{1.00} &101.76  &147.78 &0.43 \\
\bottomrule
\end{tabular}
\end{center}
\vspace{-4mm}
\end{table}

\subsection{Hierarchical time series}  \label{sec:hierarchicaltimeseries}

\header{Task} So far, our experimental results were obtained using the mean-squared error as objective function for which an analytical gradient and hessian can be easily derived. In this experiment, we apply PGBM to the problem of hierarchical time series forecasting, which is a problem where our loss function is rather complex, so that it becomes very tedious to manually calculate an analytical gradient and hessian for it: 
\begin{align} \label{eq:wmse}
L &= \sum_j^N w_j (y_j - \hat{y}_j)^2,
\end{align}
where \(w_j\) is the weight of the \(j\)-{th} time series, and \(N\) is the number of time series. In hierarchical time series, we aggregate time series across multiple levels. For example, in the case of two time series and two levels, \(N = 3\) and our loss for each series reads \(L_1 = w_1 (y_1 - \hat{y}_1)^2\), \(L_2 = w_2 (y_2 - \hat{y_2})^2\), \(L_3 = w_3 ((y_1 + y_2) - (\hat{y}_1 + \hat{y}_2))^2\) with \(w_1,w_2,w_3\) weights of each series, for example \(0.25, 0.25, 0.5\). Hence, the gradient and hessian of \(L\) with respect to the first estimate \(\hat{y}_1\) becomes \(\frac{\partial L}{\partial \hat{y}_1} = -w_1 (2y_1 - 2\hat{y}_1) - w_3(2y_1 + 2y_2 - 2\hat{y}_1 + 2\hat{y}_2) \) and \(\frac{\partial^2 L}{\partial^2 \hat{y}_1} = 2w_1 + 2w_3\). It becomes clear that deriving this result analytically becomes increasingly complex when the number of levels and the number of time series increases, which necessitates the use of autodifferentiation packages such as PyTorch if we are interested in optimizing this objective.

\header{Dataset} We use a subset of the dataset from the M5 forecasting competition \cite{makridakis_m5_2020}, in which participants were asked to provide hierarchical forecasts for Walmart retail store products. We use a single store and create forecasts for a single day. For each store, we are interested not only in accurately forecasting individual item sales, but also in optimizing the aggregate sales per day, aggregate sales per day per category and aggregate sales per day per department. Hence, we obtain four levels for our weighted loss function: 
\begin{enumerate*}[leftmargin=*,label=(\roman*)]
\item individual items, 
\item category aggregates per day, 
\item department aggregates per day, and 
\item total daily aggregates. 
\end{enumerate*}
For more details on the data and preprocessing we refer to Supplemental Materials~\ref{app:reproducibility}.

\header{Protocol} We compare against a baseline of LightGBM, NGBoost and PGBM trained with the regular mean-squared error objective. All models are trained using the same hyperparameters. We validate on the last 28 days of the training dataset and pick the number of iterations resulting in the lowest item RMSE on this validation set. After validating, we retrain on the entire dataset for the number of optimal iterations and test on a held out test set of 28 days (with the first day starting the day after the last day in the validation set). We use the Normal distribution with a tree correlation of \(\rho = \frac{\log_{10} n}{100}\) to generate our probabilistic forecasts for PGBM. 

\header{Results} We evaluate our model on RMSE and CRPS for each aggregation and the results are displayed in Table~\ref{tab:htimeseries}. We observe that using the weighted MSE that incorporates our four levels of aggregation results in a similar point forecast score for individual items, but in a much better forecast for the aggregations -- differences up to 10\% compared to the second-best point performance of PGBM are observed. Secondly, we see that the gain using the weighted MSE increases at hierarchically higher aggregation levels such as `total by date'. This is important, as this implies that we are able to generate item-level forecasts that are more consistent with higher-level aggregates. Finally, we observe that item-level CRPS is worse in the weighted MSE setting compared to the regular MSE setting, whereas our probabilistic estimate for higher aggregations improves up to 300\% when using the weighted MSE. This can be expected: in the MSE setting, each individual item forecast `does not consider' aggregates in the category or department, whereas in the weighted MSE setting, item forecasts are optimized to also consider the impact on the overall aggregations. All in all, this experiment demonstrates the benefit of our implementation: we can optimize over more complex loss functions, thereby enabling probabilistic forecasts of more complex problems such as hierarchical time series problems. 

\begin{table}
\captionsetup{font=small}
\tabcolsep=0.11cm
\footnotesize
\caption{Point (RMSE) and probabilistic (CRPS) forecasting performance for the M5 dataset across aggregations when using MSE or weighted MSE (only for PGBM) as training objective. Lower is better, best results are indicated in bold.}
\label{tab:htimeseries}
\begin{center}
\begin{tabular}{l c c c c | c c c}
\toprule 
 &\multicolumn{4}{c}{RMSE} &\multicolumn{3}{c}{CRPS} \\
 \cmidrule(r){2-5} \cmidrule(l){6-8}
 &\multicolumn{2}{c}{PGBM} &NGBoost &LightGBM & \multicolumn{2}{c}{PGBM} & NGBoost \\
\textit{objective}& \textit{MSE} & \textit{wMSE} & \textit{MSE} &\textit{MSE} & \textit{MSE} & \textit{wMSE} & \textit{MSE} \\ 
\midrule
Individual\textsuperscript{1} &\textbf{2.00} &\textbf{2.00} &2.01 &\textbf{2.00} &\textbf{0.77} &0.93 &0.78 \\
Category\textsuperscript{2} &67.9 &\textbf{67.6} &77.4 &70.0 &72.6 &\textbf{40.4} &82.0 \\
Department\textsuperscript{3}   &108 &\textbf{101} &129 &111 &160 &\textbf{59} &184 \\
Total\textsuperscript{4} &213 &\textbf{190} &276 &225 &472 &\textbf{136} &560 \\
\bottomrule
\noalign{\vskip 1mm} 
\multicolumn{8}{l}{1. n=85,372 2. n=196 3. n=84 4. n=28.}
\end{tabular}
\end{center}
\vspace{-4mm}
\end{table}


\section{Related work} 
\label{sec:related}
Traditional forecasting methods such as ARIMA \citep{box_distribution_1970} allow for probabilistic forecasts through specification of confidence intervals \citep{hyndman_forecasting_2018}. However, creating a confidence interval in these methods often requires assuming normality of the distribution of the target variable or its residuals. Generalized Additive Models for Shape, Scale and Location (GAMLSS) \citep{rigby_generalized_2005} is a framework that allows for a more flexible choice of distribution of the target variable in probabilistic regression problems. A disadvantage is that the model needs to be pre-specified, limiting flexibility of this method. Prophet \citep{taylor_forecasting_2018} is a more recent example of generalized additive models applied to the probabilistic forecasting setting. However, Prophet has been shown to underperform other contemporary probabilistic forecasting methods \citep{alexandrov_gluonts_2020, sen_think_2019} and to have difficulties scaling to very large datasets \citep{sen_think_2019}. Other Bayesian methods exhibiting similar issues include Bayesian Additive Regression Trees (BART) \citep{chipman_bart_2010}, which requires computationally expensive sampling techniques to obtain probabilistic predictions.

\acp{GBM} \citep{friedman_greedy_2001} are widely used for regression problems such as forecasting~\citep{chen_xgboost_2016}. Popular \ac{GBM} implementations such as LightGBM \citep{ke_lightgbm_2017} or xgboost \citep{chen_xgboost_2016} have won various machine learning competitions \citep{chen_xgboost_2016}. The winning solution of the accuracy track of the recent M5 forecasting competition was based on a set of LightGBM models, and 4 out of the top-5 solutions used LightGBM models in their solutions \cite{makridakis_m5_2020}. However, \acp{GBM} are not naturally equipped to provide probabilistic forecasts as these models return point predictions, requiring multiple models when a practitioner desires a range of predictions. For example, the uncertainty track of the M5 forecasting competition required contestants to provide a set of quantiles for a hierarchical time series problem. The winning solution was based on 126 (!) separate LightGBM models, one for each requested quantile and time series aggregation level \cite{makridakis_m5_2020-1}. To address these limitations, NGBoost~\cite{duan_ngboost_2020} allows for probabilistic regression with a single \ac{GBM} by using the natural gradient to boost multiple parameters of a pre-specified distribution. Compared to NGBoost, our method \ac{PGBM} does not require a natural gradient; it can achieve better or on-par predictive uncertainty estimates, without sacrificing performance on the point forecast of the same model as does NGBoost. \citet{taieb_probabilistic_2015} also propose boosted additive models for probabilistic forecasting. Our work is different in that we use GBMs with stochastic leaf weights to estimate the conditional mean and variance simultaneously for each estimator. \citet{gouk_stochastic_2019} present a method for incrementally constructing decision trees using stochastic gradient information. Our method is different in that (i) we focus on the general case of probabilistic regression instead of incremental online learning of trees and (ii) we obtain our stochastic estimates by approximating the ratio of the gradient and hessian. 

Outside of the GBM context, decision trees have also been used for probabilistic regression problems in Quantile Regression Forests (QRF) \cite{meinshausen_quantile_2006}. However, this method requires storing all observations when computing leaf weights of a decision tree, which makes this method less suitable for large datasets. In addition, GBMs commonly outperform random forests on regression tasks, making the former a better choice when performance is a key consideration. 

Contemporary large-scale probabilistic forecasting techniques often leverage the power of neural networks, such as DeepAR \citep{salinas_deepar_2019} or Transformer-based models \cite{lim_temporal_2019, li_enhancing_2019}. However, in practice GBMs still seem the technique of choice---only one out of the top-5 solutions in the M5 uncertainty forecasting competition used a neural network method as its primary probabilistic forecasting tool.

In summary, we contribute the following on top of the related work discussed above: (i) PGBM is a single-parameter boosting method that achieves state-of-the-art point and probabilistic estimates using a single model, (ii) PGBM allows choosing an output distribution after training, which means the probabilistic forecast can be optimized after training, (iii) our implementation allows training of larger datasets up to several orders of magnitude faster than the existing state-of-the-art, and (iv) our implementation allows using complex differentiable loss functions, which removes the need to calculate an analytical gradient, thereby opening up a wider set of problems that can be effectively addressed.


\section{Conclusion}
In this work we introduced Probabilistic Gradient Boosting Machines (PGBM), a method for probabilistic regression using gradient boosting machines. PGBM creates probabilistic estimates by using stochastic tree leaf weights based on sample statistics. By combining the learned weights for each subsequent tree, PGBM learns a mean and variance for samples in a dataset which can be used to sample from an arbitrary distribution of choice. We demonstrated that PGBM provides state-of-the-art probabilistic regression results across a range of small to large datasets. Benefits of PGBM compared to existing work are that 
\begin{enumerate*}[leftmargin=*,label=(\roman*)]
\item PGBM is a single-parameter boosting method that optimizes a point regression but achieves state-of-the-art probabilistic estimates using the same model, 
\item PGBM enables the choice of an output distribution after training, which means practitioners can optimize the choice of distribution without requiring retraining of the model, 
\item our implementation allows training of larger datasets up to several orders of magnitude faster than the existing state-of-the-art, and 
\item our implementation in PyTorch allows using complex differentiable loss functions which removes the need to calculate an analytical gradient as is common in existing gradient boosting packages. 
\end{enumerate*}
We demonstrated the latter benefit for the problem of hierarchical time series forecasting, where we observed up to 10\% improvement in point performance and up to 300\% improvement in probabilistic forecasting performance. 

Limitations of our work are that PGBM only learns the mean and variance in a tree, which limits the choice of output distribution. However, we observed no negative performance effects in our experiments thereof. 

In the future, we intend to further work on the theoretical error bounds of PGBM. Under mild assumptions, sample statistics in each leaf of each tree appropriately represent the true statistics of the samples in each leaf provided a sufficient number of samples. However, we have yet to determine appropriate theoretical error bounds on the final estimated statistics when considering the simplifications we make during decision tree learning, such as the greedy approximate split finding, using a limited number of tree leaves, our approximation to the stochastic leaf weights, keeping decision points constant and treating the correlation between subsequent trees as a constant across samples and trees. Regarding the latter, we also expect that the probabilistic estimate can be further improved by using a better approximation to the tree correlations instead of our choice of keeping it fixed across trees and samples.

\bibliographystyle{ACM-Reference-Format}
\bibliography{paper}
\clearpage
\appendix

\section{Derivation of stochastic leaf weights} \label{app:stweights}
\subsection{Expectation}
We approximate the mean in each leaf by using a second-order Taylor approximation of the expectation of a function of the two random variables ($g$, $h$) around the point \( \boldsymbol{a} = (\overline{g}, \overline{h})\):
\begin{align} 
E[f(g, h)] &= E[f(\boldsymbol{a}) + f'_g(\boldsymbol{a})(g - \overline{g}) + f'_h(\boldsymbol{a})(h - \overline{h}) \nonumber \\
		&\quad + \frac{1}{2} f''_{gg}(\boldsymbol{a})(g - \overline{g})^2 + f''_{gh}(\boldsymbol{a})(g - \overline{g})(h - \overline{h}) \nonumber  \\
		&\quad + \frac{1}{2} f''_{hh}(\boldsymbol{a})(h - \overline{h})^2 + H]
\end{align}
with \(H\) denoting the higher-order terms that we drop for our estimate. Using the laws of expecations we then obtain:
\begin{align} 
E[f(g, h)] &\approx E[f(\boldsymbol{a})] + E[f'_g(\boldsymbol{a})(g - \overline{g})] + E[f'_h(\boldsymbol{a})(h - \overline{h})] \nonumber \\
		& \quad + E[\frac{1}{2} f''_{gg}(\boldsymbol{a})(g - \overline{g})^2] + E[f''_{gh}(\boldsymbol{a})(g - \overline{g})(h - \overline{h})] \nonumber  \\
		& \quad + E[\frac{1}{2} f''_{hh}(\boldsymbol{a})(h - \overline{h})^2] \\
		&= E[f(\boldsymbol{a})] + f'_g(\boldsymbol{a}) \underbrace{E[(g - \overline{g})]}_{0} + f'_h(\boldsymbol{a}) \underbrace{E[(h - \overline{h})]}_{0} \nonumber \\
		& \quad + \frac{1}{2} f''_{gg}(\boldsymbol{a}) \underbrace{E[(g - \overline{g})^2]}_{\sigma^2_g} + f''_{gh}(\boldsymbol{a}) \underbrace{E[(g - \overline{g})(h - \overline{h})]}_{\sigma^2_{gh}} \nonumber  \\
		& \quad + \frac{1}{2} f''_{hh}(\boldsymbol{a}) \underbrace{E[(h - \overline{h})^2]}_{\sigma^2_g} \\ 
		&= E[f(\boldsymbol{a})] + \frac{1}{2} f''_{gg}(\boldsymbol{a})\sigma^2_g  + f''_{gh}(\boldsymbol{a})\sigma^2_{gh} \nonumber \\
		&\quad +  \frac{1}{2} f''_{hh}(\boldsymbol{a})\sigma^2_h,
\end{align}
with \(\sigma^2_{gh}\) denoting the covariance of the gradient and the hessian. 
For a function \(f(g, h) = \frac{g}{h}\), we have:
\begin{align*} \label{eq:derivatives}
f'_{g} &= h^{-1} \\
f'_{h} &= -gh^{-2} \\
f''_{gg} &= 0 \\
f''_{gh} &= -h^{-2} \\
f''_{hh} &= 2gh^{-3}.
\end{align*}
Substituting and using \( \boldsymbol{a} = (\overline{g}, \overline{h} ) \), \(f(\boldsymbol{a}) = \frac{\overline{g}}{\overline{h}}\):
\begin{align}
E[f(g, h)]  &\approx E[f(\boldsymbol{a})] -h^{-2}(\boldsymbol{a})\sigma^2_{gh} +  gh^{-3}(\boldsymbol{a})\sigma^2_h \\
		  &= \frac{\overline{g}}{\overline{h}} - \frac{\sigma^2_{gh}}{\overline{h}^2} +  \frac{\overline{g} \sigma^2_h}{\overline{h}^3}.
\end{align}
Finally, we can include the regularization constant \(\overline{\lambda}\) to arrive at the final estimate of the expectation for the leaf weight \(w_j\). This constant only affects the mean of the random variable \(h\), therefore we can safely add it to the terms containing \(\overline{h}\):
\begin{align} \label{eq:approxmean}
E\left[\frac{\overline{g}}{(\overline{h} + \overline{\lambda})}\right]  &\approx \frac{\overline{g}}{(\overline{h} + \overline{\lambda})} - \frac{\sigma^2_{gh}}{(\overline{h}  + \overline{\lambda})^2} +  \frac{\overline{g} \sigma^2_h}{(\overline{h} +  \overline{\lambda})^3}.
\end{align}
Note that we can obtain the first-order Taylor approximation of the mean by dropping the last two terms of Eq. \eqref{eq:approxmean}:
\begin{align}
E\left[\frac{\overline{g}}{(\overline{h} + \overline{\lambda})}\right]  &\approx \frac{\overline{g}}{(\overline{h} + \overline{\lambda})}.
\end{align}

\subsection{Variance}
For the variance, we start with the definition of variance for a function \(f(g, h)\):
\begin{align}
V \left[f(g, h)\right] &= E\left[\left(f(g, h) - E[f(g, h)]\right)^2\right].
\end{align}
We perform a first-order Taylor expansion of \(f(g, h)\) around the point \( \boldsymbol{a} = (\overline{g}, \overline{h})\) and we substitute the first-order approximation of the mean:
\begin{align}
V \left[f(g, h)\right] &\approx E\Big[ \Big( f(\boldsymbol{a}) + f'_g(\boldsymbol{a})(g - \overline{g}) + f'_h(\boldsymbol{a})(h - \overline{h}) \nonumber \\
		&\quad - E[f(\boldsymbol{a})]\Big)^2 \Big] \\
		&= E[ (f'_g(\boldsymbol{a})(g - \overline{g}) + f'_h(\boldsymbol{a})(h - \overline{h}))^2 ] \\
		&= E[ f^{'2}_g(\boldsymbol{a})(g - \overline{g})^2 + f^{'2}_h(\boldsymbol{a})(h - \overline{h})^2 \nonumber \\
		& \quad + 2 f'_g (\boldsymbol{a})(g - \overline{g})f'_h  (\boldsymbol{a})(h - \overline{h}) ] \\
		&= f^{'2}_g (\boldsymbol{a}) E[(g - \overline{g})^2] + f^{'2}_h(\boldsymbol{a})E[(h - \overline{h})^2] \nonumber \\ 
		& \quad + 2 f'_g (\boldsymbol{a}) f'_h  (\boldsymbol{a}) E[(g - \overline{g})(h - \overline{h}) ] \\
		&= \overline{h}^{-2} \sigma^2_g + \overline{g}^2\overline{h}^{-4} \sigma^2_h - 2 \overline{g}\overline{h}^{-3} \sigma^2_{gh} \\
		&= \frac{\sigma^2_g}{\overline{h}^2} + \frac{\overline{g}^2 \sigma_h^2}{\overline{h}^{4}} - 2 \frac{\overline{g}\sigma^2_{gh}}{\overline{h}^{3}}.
\end{align}
Finally, including the regularization constant \(\overline{\lambda}\) we obtain:
\begin{equation}
V\left[\frac{\overline{g}}{(\overline{h} + \overline{\lambda})}\right] \approx  \frac{\sigma^2_g}{(\overline{h} + \overline{\lambda})^2} + \frac{\overline{g}^2 \sigma_h^2}{(\overline{h} + \overline{\lambda})^{4}} - 2 \frac{\overline{g}\sigma^2_{gh}}{(\overline{h} + \overline{\lambda})^{3}}.
\end{equation}

\clearpage
\onecolumn

\section{Reproducibility} \label{app:reproducibility}
We report absolute scores and dataset statistics for the UCI benchmark in Table~\ref{tab:ucibenchmark}. An overview of the key hyperparameters for each method for both experiments is given in Table~\ref{tab:exphyperparam}, and absolute timings for the timings of Table~\ref{tab:timing} in Table~\ref{tab:abstiming}. An overview of the M5 dataset is given in Table~\ref{tab:exp2dataset}. We refer to our code at \url{https://github.com/elephaint/pgbm} for further details, such as the features of the M5 dataset, which mainly comprise lagged target variables, time indicators (e.g., day-of-week), event indicators (e.g., holidays) and item indicators.

\begin{table}[h!]
\caption{Results for probabilistic (CRPS) and point (RMSE) performance for each dataset. We report mean metrics over all folds per method and indicate the standard deviation in brackets. Lower is better.}
\label{tab:ucibenchmark}
\begin{center}
\begin{tabular}{l c c c c c c c c}
\toprule 
	& & & & \multicolumn{2}{ c }{CRPS} &  \multicolumn{3}{ c }{RMSE} \\
 \cmidrule(r){5-6} \cmidrule(r){7-9}
Dataset  & folds &samples &features & PGBM & NGBoost & PGBM & NGBoost & LightGBM \\
\midrule
\texttt{yacht} 	&	20	&	308	&	6	&0.22 (0.070)	&0.32 (0.104)	&0.63 (0.213)	&0.75 (0.297)	&0.64 (0.281) \\
\texttt{boston} 	&	20	&	506	&	13	&1.61 (0.201)	&1.73 (0.236)	&3.05 (0.507)	&3.31 (0.661)	&3.11 (0.675) \\
\texttt{energy} 	&	20	&	768	&	8	&0.21 (0.034)	&0.25 (0.022)	&0.35 (0.062)	&0.49 (0.055)	&0.29 (0.075) \\
\texttt{concrete} 	&	20	&	1,030	&	8	&2.06 (0.335)	&2.95 (0.326)	&3.97 (0.759)	&5.50 (0.642)	&3.80 (0.762) \\
\texttt{wine} 	&	20	&	1,599	&	11	&0.33 (0.034)	&0.34 (0.024)	&0.60 (0.054)	&0.62 (0.043)	&0.60 (0.050) \\
\texttt{kin8nm} 	&	20	&	8,192	&	8	&0.07 (0.002)	&0.10 (0.002)	&0.13 (0.005)	&0.17 (0.003)	&0.11 (0.003) \\
\texttt{power} 	&	20	&	9,568	& 4 	&1.81 (0.053)	&2.01 (0.120)	&3.35 (0.153)	&3.70 (0.222)	&3.20 (0.140) \\
\texttt{naval} 	&	20	&	11,934	&	14	&0.00 (0.000)	&0.00 (0.000)	&0.00 (0.000)	&0.00 (0.000)	&0.00 (0.000) \\
\texttt{protein} 	&	20	&	45,730	& 9		&2.19 (0.030)	&2.44 (0.038)	&3.98 (0.056)	&4.50 (0.059)	&3.82 (0.058) \\
\texttt{msd} 	&	1	&	515,345	& 90		&4.78	&4.75	&9.09	&9.16	&9.11 \\
\texttt{higgs} 	& 1 		& 10,500,000 & 28 	& 0.253 					& 0.238					& 0.418					&0.419 					&0.414 \\
\bottomrule
\end{tabular}
\end{center}
\end{table}

\begin{table}[h!]
\caption{Key hyperparameters for the UCI benchmark and hierarchical time series experiment.}
\label{tab:exphyperparam}
\begin{center}
\begin{tabular}{l c c c c c c}
\toprule
 & \multicolumn{3}{ c }{UCI benchmark} &  \multicolumn{3}{ c }{Hierarchical time series} \\ 
 \cmidrule(r){2-4} \cmidrule(r){5-7}
   & PGBM &NGBoost &LightGBM &PGBM &LightGBM &NGBoost \\
\midrule
min\_split\_gain &0 &0 &0 &0 &0 &0\\
          min\_data\_in\_leaf  &1 &1 &1 &1 &1 &1\\
          max\_bin &64 &n.a. &64 &1024 &1024 &n.a.\\
          max\_leaves &16 &n.a. &16 &64 &64 &64\\
		max\_depth &-1 &3 &-1 &-1 &-1 &-1\\
          learning\_rate &0.1 &0.01 &0.1 &0.1 &0.1 &0.1\\
          n\_estimators &2000 &2000 &2000 &1000 &1000 &1000\\
          feature\_fraction &1.0 &1.0 &1.0 &0.7 &0.7 &0.7\\
          bagging\_fraction &1.0 &1.0 &1.0 &0.7 &0.7 &0.7\\
          seed &1 &1 &1 &1 &1 &1\\
          lambda &1.0& n.a. &1.0 &1.0 &1.0 &n.a.\\
          early\_stopping\_rounds &n.a. &n.a. &n.a. &20 &20 &20\\
\bottomrule
\end{tabular}
\end{center}
\end{table}

\begin{table}[h!]
\parbox{.45\linewidth}{
	\centering
	\caption{Average time in seconds for running 2,000 iterations for each dataset on the UCI benchmark datasets. For \texttt{msd} and \texttt{higgs}, a bagging fraction of 0.1 was used.}
	\label{tab:abstiming}
	\begin{tabular}{l c c c | c}
	\toprule 
			& \multicolumn{3}{c}{Probabilistic forecast}	&Point forecast  \\
	Dataset  & \textit{PGBM-gpu} &\textit{PGBM-cpu} & \textit{NGBoost} & \textit{LightGBM} \\
	\midrule
	\texttt{wine} (n=1,599)  &	100 &	41 &20 &1 \\
	\texttt{naval} (n=12k) &	103 &	167 &108 & 2 \\
	\texttt{protein} (n=46k) &	115 &	355 &389 & 2 \\
	\texttt{msd} (n=515k) & 136 &3,645 &6,628 &12 \\
	\texttt{higgs} (n=10,5M) & 316  &32,200  &46,744 &135 \\
	\bottomrule
	\end{tabular}
	}
\hfill
\parbox{.45\linewidth}{
	\centering
	\caption{M5 dataset description.}
	\label{tab:exp2dataset}
	\begin{tabular}{l c c}
	\toprule 
   	& & \texttt{M5} \\
	\midrule
	time series & \# & 3,049 \\
	time series description & &item product sales \\
	target & & \(\mathbb{R}^+\) \\
	train samples & \# & 2,415,359 \\
	validation samples & \# & 85,372 \\
	test samples & \# & 85,372 \\
	time step & \(t\) & day \\
	features & \# & 48 \\
	\bottomrule
	\end{tabular}
	}
\end{table}

\end{document}